\documentclass{article}

    \PassOptionsToPackage{numbers, compress}{natbib}

\usepackage[preprint]{neurips_2024}

\usepackage[utf8]{inputenc} 
\usepackage[T1]{fontenc}    
\usepackage{hyperref}       
\usepackage{url}            
\usepackage{booktabs}       
\usepackage{amsfonts}       
\usepackage{nicefrac}       
\usepackage{microtype}      
\usepackage{xcolor}         
\usepackage{amsmath}        
\usepackage{graphicx}       
\usepackage{multirow}       
\usepackage{enumitem}       
\usepackage{lineno}         
\usepackage[export]{adjustbox}  
\usepackage[normalem]{ulem}
\usepackage{amssymb}        
\usepackage{pifont}         
\usepackage{tabularx}       
\usepackage{subcaption}     
\usepackage{float}          
\usepackage{footmisc} 

\newcommand{\cmark}{\ding{51}}%
\newcommand{\xmark}{\ding{55}}%
\useunder{\uline}{\ul}{}

\title{RICASSO: Reinforced Imbalance Learning with Class-Aware Self-Supervised Outliers Exposure}

\author{
  Xuan Zhang, Sin Chee Chin, Tingxuan Gao, Wenming Yang\thanks{Corresponding author}\\
  Shenzhen International Graduate School, Tsinghua University, Shenzhen, China \\
  \texttt{\{zhangxua22, chenzxz22, gaotx22\}@mails.tsinghua.edu.cn}, \\
  \texttt{yang.wenming@sz.tsinghua.edu.cn} 
}

\begin{document}

\maketitle

\begin{abstract}

In real-world scenarios, deep learning models often face challenges from both imbalanced (long-tailed) and out-of-distribution (OOD) data. However, existing joint methods rely on real OOD data, which leads to unnecessary trade-offs. 
In contrast, our research shows that data mixing, a potent augmentation technique for long-tailed recognition, can generate pseudo-OOD data that exhibit the features of both in-distribution (ID) data and OOD data. Therefore, by using mixed data instead of real OOD data, we can address long-tailed recognition and OOD detection holistically.
We propose a unified framework called \textbf{R}einforced \textbf{I}mbalance Learning with \textbf{C}lass-\textbf{A}ware \textbf{S}elf-\textbf{S}upervised \textbf{O}utliers Exposure (RICASSO), where "self-supervised" denotes that we only use ID data for outlier exposure. RICASSO includes three main strategies:
\textbf{Norm-Odd-Duality-Based Outlier Exposure}: Uses mixed data as pseudo-OOD data, enabling simultaneous ID data rebalancing and outlier exposure through a single loss function.
\textbf{Ambiguity-Aware Logits Adjustment}: Utilizes the ambiguity of ID data to adaptively recalibrate logits.
\textbf{Contrastive Boundary-Center Learning}: Combines Virtual Boundary Learning and Dual-Entropy Center Learning to use mixed data for better feature separation and clustering, with Representation Consistency Learning for robustness.
Extensive experiments demonstrate that RICASSO achieves state-of-the-art performance in long-tailed recognition and significantly improves OOD detection compared to our baseline (27\% improvement in AUROC and 61\% reduction in FPR on the iNaturalist2018 dataset). On iNaturalist2018, we even outperforms methods using real OOD data. The code will be made public soon.

\end{abstract}

\section{Introduction}

Despite the significant advancements in deep learning, it still faces challenges when adapting to real-world scenarios, where data imbalance and out-of-distribution (OOD) data coexist. Although numerous effective strategies have been developed to individually address long-tailed recognition\cite{cRT2019, Remix2020, LFME2020, LogitAdj2021, LGLA2023} or OOD detection \cite{ MD2018, ODIN2018, EBO2020, MSP2022, OOD2prior2023}, they often fall short when confronted with these issues simultaneously.
Data imbalance, also known as the long-tailed distribution, causes the model to incorrectly predict underrepresented tail classes as well-represented head classes.
Similarly, deep learning models are prone to make erroneous predictions for unknown or OOD data, potentially leading to significant errors in application \cite{EAT2024}.
These factors introduce new problems when data imbalance and OOD data appear concurrently:
Firstly, the model tends to bias towards the well-represented head classes in imbalanced settings, leading to misclassification of OOD samples as head classes.
On the other hand, given that both tail and OOD samples are underrepresented, the model is prone to mistaking tail classes samples for OOD.
Therefore, a methodology that simultaneously tackles long-tailed recognition and OOD detection is necessitated.

\begin{figure}[H]
    \centering
    \includegraphics[height=0.5\linewidth]{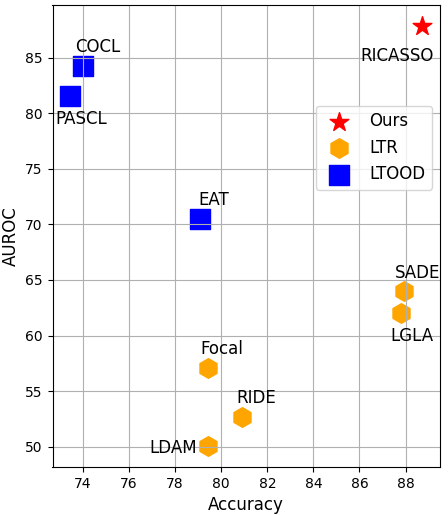}
    \caption{Ours vs. other long-tailed recognition (LTR) and long-tailed OOD (LTOOD) detection methods on CIFAR-10, IR100. Without the need for real OOD data, our RICASSO outperforms all of the other methods on the iNaturalist2018 \cite{iNature2018} dataset.}
    \label{fig:result}
\end{figure}


Current joint methods for long-tailed recognition and OOD detection often incorporate traditional OOD methods into the long-tailed recognition framework \cite{COCL2024, OpenSampOOD2022, EAT2024, BalEnergy2023, PASCL2022}.
COCL \cite{COCL2024} integrates a contrastive learning method to explicitly distance tail samples from OOD samples, while OpenSampling \cite{OpenSampOOD2022} and EAT \cite{EAT2024} mix the in-distribution (ID) samples with the OOD data, to rebalance the long-tailed distribution.
However, the introduction of real OOD data may result in unnecessary trade-offs, a fact also supported by the work of \cite{PASCL2022, UD2017, InvIOD2023}. 
In contrast to PASCL's approach of mitigating the trade-off through two-stage training \cite{PASCL2022}, our method aims to \textbf{eliminate the need for real OOD data}. 
Therefore, we propose a unified framework named \textbf{R}einforced \textbf{I}mbalance Learning with \textbf{C}lass-\textbf{A}ware \textbf{S}elf-\textbf{S}upervised \textbf{O}utliers Exposure (RICASSO), which addresses long-tailed recognition and OOD detection issues simultaneously. The term "Self-Supervised" indicates that our RICASSO conducts outlier exposure without using real OOD data.

\nolinenumbers

\nolinenumbers
The first hurdle of RICASSO is to find a suitable substitute for real OOD data. We discovered that mixed data, which is effective for long-tailed recognition, exhibits features of both ID and OOD data. We refer to this property as \textit{Norm-Odd Duality}. Utilizing this, we propose \textbf{Norm-Odd-Duality-Based Outlier Exposure} (NOD) to combine long-tailed recognition and outlier exposure in a single loss function. Additionally, we sample the imbalanced dataset in an anti-long-tailed manner and mix it with the original long-tailed data. This results in frequent head and tail class pairs, compelling the model to distinguish between them and achieve debiasing.

RICASSO's second challenge is how to integrate long-tailed recognition with OOD detection, to prevent the trade-off between them. Conventional logits adjustment methods only use a static class-wise prior \cite{LDAM2019, LogitAdj2021, LGLA2023, RIDE2020, SADE2022}, which has two limitations: it lacks subtle attention for each sample and cannot adapt as the network iterates. Since both tail class data and OOD data are underrepresented, we propose using the outlier score as supplemental information for the prior. Our method, \textbf{Ambiguity-Aware Logits Adjustment} (AALA), uses the outlier score to recalibrate the prior, providing detailed attention for each sample and adaptability throughout the learning process.

The third obstacle in RICASSO lies in leveraging mixed data for contrastive learning, essential for OOD detection. Existing methods rely on real OOD data \cite{COCL2024, PASCL2022}, which has proven inferior  \cite{PASCL2022, UD2017, InvIOD2023}. Additionally, their clustering processes do not adapt to varying sample difficulties in OOD detection.
Given that NOD ensures mixed data features are positioned between their source classes, we use this as \textbf{pseudo boundaries} for feature clustering, called \textbf{Virtual Boundary Learning}. To enhance feature clustering, we introduce \textbf{Dual-Entropy Center Learning}, which uses dual entropy to learn more compact feature distributions for each class. Lastly, inspired by SimCLR \cite{SimCLR2020}, we integrate \textbf{Representation Consistency Learning} to extract consistent information from mixed data obtained through different mixing techniques, improving the model’s robustness.

Our principal contributions are as follows: 
\noindent
\begin{itemize}[noitemsep, left=0pt]
    \item We are the first to uncover the Norm-Odd Duality of mixed ID samples and propose a unified framework for the joint tasks of long-tailed recognition and OOD detection. Our RICASSO eliminates the need for real OOD data, thus mitigating the trade-off introduced by it.
    \item We incorporate the outlier score as a measure of tailness into the conventional logits adjustment method. This allows the method to adapt during the training process and to provide more subtle attention for each sample.
    \item We further explore the role of mixed data as a virtual boundary in feature space, which is utilized to enhance representation learning. 
    \item Extensive experiments prove that RICASSO demonstrates the best comprehensive performance on the tasks of long-tailed recognition and OOD detection. Not only do we achieve SOTA in long-tailed recognition, we also improve OOD detection by a substantial margin (\textbf{27\%} improvement of AUROC and \textbf{61\%} reduction of FPR on iNaturalist2018). On iNaturalist2018 \cite{iNature2018}, our RICASSO even outperforms all of the long-tailed OOD methods that \textbf{utilize real OOD data} (as shown in Figure \ref{fig:result}).
\end{itemize}

\section{Related Work}
\label{sec:related_work}

\paragraph{Basic solutions for long-tailed recognition}
Current approaches to long-tailed recognition predominantly concentrate on two dimensions: data and algorithms.
\textbf{Data-focused strategies} aim to rebalance the data distribution through either resampling \cite{SMOTE2002, BorderlineSMOTE2005, DCL2019, LRSMOTE2020, BalMS2020, LOCE2021} or  data augmentation\cite{Remix2020, GLMC2023, RankMix2023, BEM2024}. 
\textbf{On the algorithmic side}, approaches include category-sensitive learning and transfer learning. 
	Most \underline{category-sensitive learning} methods adjust the training loss values for each category, utilizing techniques such as reweighting\cite{Focal2017, GHM2019, EffectiveNumLT2019, AREA2023, LDAM2019, EqLossGrad2023, LiVT2023} or remargining\cite{LDAM2019, BalMS2020, LGLA2023, LiVT2023, LogitAdj2021, LOCE2021}. 
	\underline{Transfer learning}, meanwhile, leverages knowledge from one area to strengthen model training in another, incorporating three main tactics: Two-stage training\cite{cRT2019, MiSLAS2021}, which starts with training the model on an imbalanced dataset and then retraining the classifier on a balanced dataset. Model ensemble\cite{LFME2020, BBN2020, RIDE2020, SADE2022, ResLT2023, LGLA2023, ResLT2023, MDCS2023, NCL++2024}, which merges insights from various experts with distinct capabilities to produce a more balanced output. Finally, head-to-tail transfer\cite{FTL2019, TransH-T2023, BEM2024} aims to leverage knowledge of head classes to improve performance on tail classes.

\paragraph{Out-of-Distribution Detection}
Existing methodologies for OOD detection in classification can be catagoried into two main approaches. 
\textbf{Post-hoc} methods use outputs like features or logits to detect OOD samples, without interfering with the training process \cite{MSP2016,ODIN2018,kNN-OOD2022,ViM2022,NNGUIDE2023}. However, they struggle with recognizing inliers and outliers effectively. 
In contrast, \textbf{outlier exposure} adds a regularization term to the training objective to help recognize OOD samples\cite{ENERGYOE2020,ATOM2021,FSOOD1023}. Yet, this could significantly harm the training objectives \cite{UD2017} and impose unnecessary constraints on the model for recognizing specific types of OOD data. 

\paragraph{Long-tailed Recognition with Out-of-Distribution Data}
Some works\textbf{ explicitly separate OOD data from tail classes}. COCL \cite{COCL2024} uses a margin-based learning approach to segregate OOD data from long-tailed data, while PASCL \cite{PASCL2022} addresses the challenge through contrastive learning.
The Balanced Energy Regularization Loss (BERL) \cite{BalEnergy2023} incorporates a class-specific prior into the standard energy function for OOD detection, aiming to reduce the model's bias towards head classes. Class Prior \cite{CLASSPRIOR2023} also make use of prior, adapting the conventional outlier exposure method to promote prior prediction instead of uniformly distributed prediction for OOD samples.
Additionally, some methods \textbf{leverage OOD data to enhance the representation of long-tailed data}. Open-Sampling \cite{OpenSampOOD2022} uses a dynamic sampling strategy to select OOD data similar to ID classes, enriching underrepresented tail classes. COLT \cite{COLT2022} employs a complementary online sampling approach within self-supervised learning framework. EAT \cite{EAT2024} introduces multiple absent classes to better represent OOD data and uses OOD data for image-wise augmentation of tail classes, enhancing their diversity.

\section{Methodology}
\label{sec:methodology}

\subsection{Preliminaries}
\label{sec:preliminaries}
The aim of long-tailed recognition is to obtain a well-balanced representation from unevenly distributed data. The long-tailed ID training set, denoted as \(\mathcal{D}_{\text{in}} = \{x_i, y_i\}_{i=1}^{N}\), consists of \(N\) training samples with \(y_i\) representing the ground truth label for the image \(x_i\). 
The total number of training samples is represented by \(N=\sum_j^C n_j\), in which \(C\) is the total number of classes, \(n_j\) is the number of samples for each class. In accordance with previous studies \cite{RIDE2020, SADE2022, LGLA2023}, we define the prior as \(\phi=\{\hat{n}_c\}^C_{c=1}\), where \(\hat{n}_c = n_c/N\).  The imbalance ratio  is calculated as the maximum \(n_j\) divided by the minimum \(n_j\), i.e., \(\max(n_j)/\min(n_j)\).

\begin{figure}
    \centering
    \begin{subfigure}[c]{\linewidth}
        \centering
        \includegraphics[width=1\linewidth]{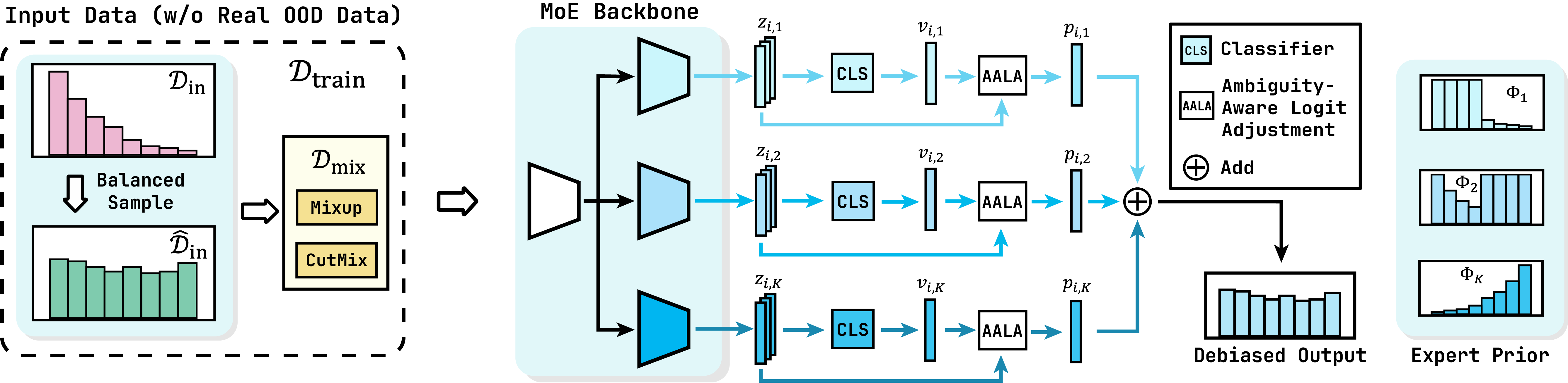}
        \caption{The pipeline of RICASSO.}
        \label{fig:pipeline}
    \end{subfigure}

    \begin{subfigure}[c]{\linewidth}
        \centering
        \includegraphics[width=1\linewidth]{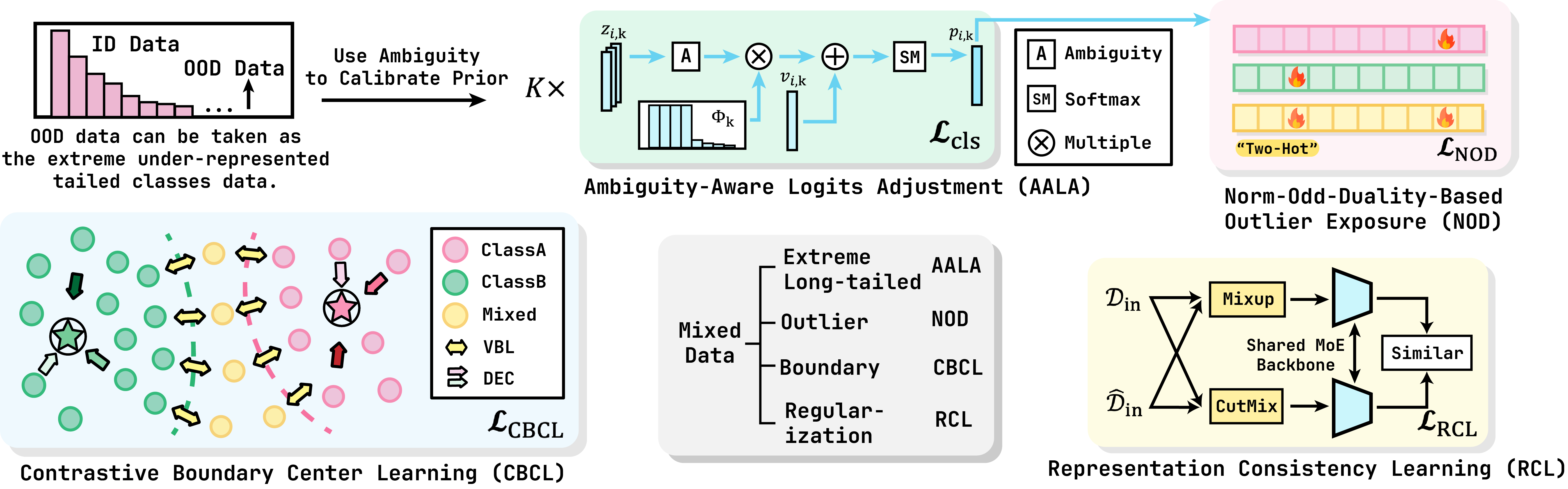}
        \caption{Each module of RICASSO.}
        \label{fig:modules}
    \end{subfigure}
    \caption{
    The overview of RICASSO. 
    (\subref{fig:pipeline}) The pipeline of RICASSO. In the data mixing phase, both Mixup\cite{Mixup2018} and Cutmix \cite{CutMix2019} is performed to generate the mixed data \(\mathcal{D}_{\text{mix}}\). Both the ID data and mixed data are used as training data. The backbone is a Mixture of Experts (MoE) network.
    (\subref{fig:modules}) The elaboration of each module. AALA (Green) takes OOD data as the extreme under-represented tailed classes data and thus uses the ambiguity of features to re-calibrate the ID priors for each expert. The calibrated logits are then fed to NOD to conduct a pseudo outlier exposure (Pink) without the need for real-OOD data. Finally, to further boost the feature clustering, the mixed data is also used for contrastive learning (Blue) and representation learning (Yellow).}
\end{figure}

Figure \ref{fig:pipeline} illustrates the overall framework of RICASSO.
As shown in the left part of the figure,  we mix the ID data \(\mathcal{D}_{\text{in}}\) and its anti-long-tailed sampled version \(\hat{\mathcal{D}}_{\text{in}}\) to get the mixed data \(\mathcal{D}_{\text{mix}}=\{x_{\text{mix}(i,j)}, y_{\text{mix}(i,j)}\}\). A mixed sample is constituted by the mixture of two source samples, which is denoted as subscript \(\text{mix}(i,j)\). Therefore, we get the whole training set \(\mathcal{D}_{\text{train}} = \{\mathcal{D}_{\text{in}}, \hat{\mathcal{D}}_{\text{in}}, \mathcal{D}_{\text{mix}}\}\). Unlike conventional outlier exposure, RICASSO does not need real-OOD data in training. 

As shown in right side of Figure \ref{fig:pipeline}, we use a Mixture of Expert (MoE) framework LGLA \cite{LGLA2023} \footnote{The LGLA code can be found at https://github.com/Tao0-0/LGLA. Also, read this paper for further explanation of the MoE network and how each expert is defined.} as the backbone for long-tailed recognition, defined as \(f_{\theta}\). 
There are \(K\) experts in total. For each \(k^{th}\) expert, we assign a group of classes for them to specialize in, denoted as \(C_k\). The last expert (the \(K^{th}\)) is an anti-long-tailed global expert, whose \(C_K\) contains all the classes, while the others are all local experts. Note that we assign different priors \(\phi_k\) for each \(k^{th}\) expert network, as shown in Equation \ref{eq:expert_prior}. \(\tau\) is a scaling factor. The different priors are also illustrated in the right part of Figure\ref{fig:pipeline}.
\begin{equation}
    \label{eq:expert_prior}
    \hat{n}_{k, j}= \left\{\begin{array}{lll}
        \hat{n}_j , & j \in C_k, k \neq K \\
        \max(\hat{n}_j), &j \notin C_k, k \neq K \\
        e^{\tau}  \hat{n}_j , &k = K \\
    \end{array}\right.
\end{equation}

For each \(k^{th}\) expert, the feature of the input image is represented as \(z_{i, k}\), and the corresponding logits is given by  \(v_{i, k}=f_\text{cls}(z_{i, k})\), where \(f_\text{cls}\) symbolizes the classifier. The \(j^{th}\) element of the logits \(v_{i, k}\) is indicated with a superscript, i..e., \(v_{i, k}^{(j)}\). The final logits are the ensemble of all the expert logits, which can be formulated as: \(v_{i} = \frac{1}{K} \Sigma_{k=1}^K v_{i, k}\). 
The probability of the current image belonging to each class is given by \(p_{i,k}\). Note that the probability is calculated using the \textbf{adjusted} logits instead of \(v_{i,k}\).

\subsection{Norm-Odd Duality based Outlier Exposure}
\label{sec:NOD}
As stated in Section \ref{sec:related_work}, conventional outlier exposure introduce a new training objective, which is producing anomalous results for anomalous samples \cite{ENERGYOE2020,ATOM2021,FSOOD1023}.
As such, the objective function of conventional outlier exposure can be formulated as:
\begin{equation}
    \label{eq:oe}
    \mathcal{L}_{\text{OE}} = \mathcal{L}_{\text{in}} + \lambda \cdot \mathcal{L}_{\text{out}}
\end{equation}
where \(\mathcal{L}_{\text{in}}\) is the ID classification loss, and \(\mathcal{L}_{\text{out}}\), as mentioned above, refers to the newly defined training objective. \(\lambda\) is the trade-off factor to control the magnitude of outlier exposure. 
However, conventional outlier exposure relies on real OOD data, which will harm the ID classification \cite{PASCL2022, UD2017}.

In contrast, we find that the mixed data exhibits characteristics of both ID and OOD data, which we term as \textit{Norm-Odd Duality}. Therefore, by replacing real OOD data with a mixed training set \(\mathcal{D}_{\text{mix}}\), we can alleviate the trade-off.

This new strategy for outlier exposure is called \textbf{N}orm-\textbf{O}dd-\textbf{D}uality-Based Outlier Exposure (NOD), providing a comprehensive perspective that considers long-tailed recognition and OOD detection as a whole. The loss function of NOD is defined in Equation \ref{eq:NOD}.
\begin{equation}
    \label{eq:NOD}
    \mathcal{L}_{\text{NOD}} = \mathbb{E}_{(x,y) \sim \mathcal{D}_{\text{train}}}[\mathcal{L}_{\text{cls}}(f_{\theta}(x), y)]
\end{equation}
In Equation \ref{eq:NOD}, \(y\) is the ground truth label for \(x\), and \(\mathcal{L}_{\text{cls}}\) is a classification loss that will be described in Section \ref{sec:AALA}. Note that for ID data, \(y\) is a one-hot array, while for mixed data, \(y\) is the weighted sum of the two source classes' label, denoted as \textit{two-hot code}.

The illustration of \textbf{N}orm-\textbf{O}dd-\textbf{D}uality-Based Outlier Exposure is depicted in the pink section of Figure \ref{fig:modules}. In NOD, ID samples are classified as a single class, while mixed samples yield predictions indicating both its source classes. This method allows NOD to conduct outlier exposure and long-tailed recognition seamlessly:
Firstly, similar to conventional outlier exposure, NOD teaches the model to predict an anomalous result (two-hot array) for anomalous samples (the mixed data). Secondly, through the two-hot prediction, NOD encourages the model to distinguish the features of the two source classes, even when they are mixed. As mentioned in Section \ref{sec:preliminaries}, we blend data from a long-tailed dataset \(\mathcal{D}_{\text{in}}\) and its re-balanced version \(\hat{\mathcal{D}}_{\text{in}}\), so the most frequently occurring mixtures are the head and tail classes. Consequently, the model is deliberately trained to differentiate between the head and tail classes, thereby debiasing its predictions.

\subsection{Ambiguity-Aware Logits Adjustment}
\label{sec:AALA}
The conventional logits adjustment method \cite{LogitAdj2021} employs class-wise prior of ID data to rebalance the biased output of the model. As shown in Equation \ref{eq:logits-adjustment}, the probability output is represented by \(p\left(v_i, k\right)\), while the prior-guided margin is denoted as \(T\left(k, j\right) = \log \left( \hat{n}_{j,k}\right)\).
\begin{equation}
\label{eq:logits-adjustment}
    p\left(v_i, k\right) = 
    \frac{\exp \left(v_{i, k}^{(y_i)}+T\left(k, y_i\right)\right)}
         {\sum_{j=1}^C \exp \left(v_{i, k}^{(j)}+T(k, j)\right)}
\end{equation}
The original prior \(\phi\) is statistically inferred from the distribution of the training data, which presents two notable limitations: First, being class-wise, the prior can't provide subtle attention for each sample. Second, the prior is static and cannot be adaptively adjusted as the network iterates.

We assume that OOD data can be considered the extremely underrepresented case of tail data, thus, the outlier score of a sample is essentially the \textit{tailness score}. Consequently, we propose \textbf{A}mbiguity-\textbf{A}ware \textbf{L}ogits \textbf{A}djustment (\textbf{AALA}) that re-calibrates the prior according to the ambiguity of the \textbf{feature} \(z_{i,k}\) of each expert, as shown in Equation \ref{eq:recalibrated-margin}.  For the sake of simplicity we omit the \(k\).
\begin{equation}
\label{eq:recalibrated-margin}
    \hat{T}(j, z_{i}) = \frac{e^ {E(z_{i})} + \sum_{n=1}^N e^{E(z_{n})}}{\sum_{n=1}^N e^{E(z_{n})}} \cdot T(j)
\end{equation}
The outlier score of the ID samples can be computed through the energy function \cite{EBO2020} as shown in Equation \ref{eq:outlier-score-energy}. 
\begin{equation}
\label{eq:outlier-score-energy}
    E(z) = - \tau \cdot \log \sum_{c=1}^C \exp \left( \dfrac{z_{c}}{\tau} \right)
\end{equation}
As this energy score is not bounded \cite{EOD2020}, we perform softmax on the energy function to normalize it with respect to the rest of the in-distribution classes. However, the normalized value is less than 1, and using it directly will significantly diminish the adjustment factor of the head classes as they have lower energy functions. To prevent this, we add 1 to the softmax score, preventing the adjustment margin \( \hat{T}(j, z_{i})\) of the head classes from becoming zero. 

\(\hat{T}(j,z_{i})\) adaptively adjust the margin of each sample based on both their prior and their OOD score. Since \(E(z_{i})\) is computed for each sample at every iteration, it also allows for a more refined adjustment tailored to the difficulty of each sample. 

Finally, we define the logits-adjustment-based classification loss \(\mathcal{L}_{\text{cls}}\) as shown in Equation \ref{eq:cls}, illustrated in the green section of Figure \ref{fig:modules}. The term \(g(k, z_{i})\) represents a re-weighting factor in accordance with the LGLA method \cite{LGLA2023}.
\begin{equation}
\label{eq:cls}
     \mathcal{L}_{\text{cls}} =
     - \sum_{k=1}^{K} g(k, z_{i,k}) 
        \log \frac
            {\exp \left(v_{i, k}^{(y_i)}+\hat{T}\left(k, y_i, z_{i,k}\right)\right)}
            {\sum_{j=1}^C \exp \left(v_{i, k}^{(j)}+\hat{T}(k, j, z_{i,k})\right)}
            , (i=1,...,N)
\end{equation}

\subsection{Contrastive Boundary-Center Learning}
\label{sec:CBCL}

\subsubsection{Virtual Boundary Learning}
\label{sec:VBL}
As outlined in Section \ref{sec:NOD}, NOD compels the mixed data to resemble both of its two source classes. Given that the final feature of the model exhibits strong semantic information, it is reasonable to assume that the feature of the mixed data, denoted as \(z_{\text{mix}{(i,j)}}\), also displays similarity to the features of both its source samples, \(z_i\) and \(z_j\). The subscript \(\text{mix}{(i,j)}\) of \(z_{\text{mix}{(i,j)}}\) indicates that the feature is derived from the mixture of samples \(x_i\) and \(x_j\).

Consequently, NOD ensures that \(z_{\text{mix}{(i,j)}}\) is positioned between the features of its two source classes, as shown in the blue section of Figure \ref{fig:modules}. 
Therefore, we propose \textbf{V}irtual \textbf{B}oundary \textbf{L}earning (\textbf{VBL}), which uses the features of mixed data as the virtual boundary to improve the separation of the ID classes, as presented in Equation \ref{eq:VBL}.
It should be noted that VBL is conducted independently within each expert, but the subscript denoting the \(k^{th}\) expert is omitted again for the sake of simplicity.

\begin{equation}
\label{eq:VBL}
    d_{-} = \mathbb{E}_{x_i \sim D_{i n} ,\space  x_j \sim \hat{D}_{i n}}
    \left( \lVert z_i - z_{\text{mix}{(i,j)}} \rVert ^ 2 _ 2 + \lVert z_j - z_{\text{mix}{(i,j)}} \rVert ^ 2 _ 2\right)
\end{equation}

VBL promotes the separation of the two source features \(z_i, z_j\) from the virtual boundary \(z_{\text{mix}{(i,j)}}\), acting as the pushing force in contrastive learning.
This results in more distinct clusters within the feature space, thereby improving OOD detection. Additionally, since the head and tail classes are often the most common pair of mixtures, VBL also helps the model to differentiate them, leading to debiasing.

\subsubsection{Dual-Entropy Center Learning}
\label{sec:DEC}
To further aid VBL, we propose \textbf{D}ual-\textbf{E}ntropy \textbf{C}enter Learning (\textbf{DEC}). This serves as the pulling force within contrastive learning, as opposed to the pushing force defined in Equation \ref{eq:VBL}.

The traditional center loss approach may lead to model bias by prioritizing the errors minimization of head classes. To address this issue, we propose incorporating hard sample mining into the center loss by assigning higher weights to difficult instances in the loss function. This integration is illustrated by the single-headed arrow (denoted as DEC in the legend) within the blue section of Figure \ref{fig:modules}. The density of the arrow's color represents the level of focus of the model.

In practice, this level of focus is calculated by a \textit{dual entropy} weight. As shown in Equation \ref{eq:dual-entropy}, dual entropy is a combination of the model's output entropy (self-entropy) and correctness (cross-entropy). This allows the center loss to place more importance on samples with lower confidence and more deviated predictions.

\begin{equation}
\label{eq:dual-entropy}
    \omega(v, y) = - \sum_{i=0}^N (v_i + y_i) \cdot \log(v_i) 
\end{equation}
Therefore, Dual-Entropy Center Learning could be defined in Equation \ref{eq:DEC}.

\begin{equation}
\label{eq:DEC}
    d_{+} = \frac{1}{2} \mathbb{E}_{x_i \sim D_{train}} \sum_{i=1}^C \omega(v_i, y_i) \cdot \lVert z_i - c_{y_i} \rVert ^ 2 _ 2
\end{equation}

Using Equation \ref{eq:VBL} and Equation \ref{eq:DEC}, we can define the loss function for Contrastive Boundary-Center Learning (CBCL) as shown in Equation \ref{eq:tritanh_loss}. 
The mathematical form of the contrastive loss function is borrowed from the work of \cite{CONSULT2024} \footnote{This reference is anonymized to comply with double-blind review requirements. 
It refers to a manuscript that the authors have yet to submit, and will be disclosed after the review process.}, and it is a more stable modification of the anchor loss.
\begin{equation}
\label{eq:tritanh_loss}
    \mathcal{L}_{CBCL} = 
    \dfrac{e^{\gamma_0 \cdot d_{+}} - e^{\gamma_1 \cdot d_{-}} + \epsilon_0}
    {e^{\gamma_0 \cdot d_{+}} + e^{\gamma_1 \cdot d_{-}} + \epsilon_1}
\end{equation}

\subsection{Representation Consistency Learning}
\label{sec:rcl}

To fully utilize the mixed data, we use them for Representation Consistency Learning (RCL) by employing similarity optimization, akin to SimCLR \cite{SimCLR2020}. 
Concretely, images augmented via MixUp \cite{Mixup2018} and CutMix \cite{CutMix2019} are fed through an encoder network followed by a projection head to derive the representations \(h_m\) and \(h_c\) respectively.
To prevent all outputs from "collapsing" to a constant value, following the approach in SimSiam \cite{SimSam2021}, we employ a stop gradient operation on both \(h_m\) and \(h_c\).  Subsequently, a prediction head processes these representations to produce outputs \(u_m\) and \(u_l\). 
We then aim to minimize their negative cosine similarity, as detailed in Equation \ref{eq:RCL}.
\begin{equation}
    \label{eq:RCL}
    \mathcal{L}_{\text{RCL}} = - \hat{u}_m \cdot \hat{h}_c - \hat{u}_c \cdot \hat{h}_m
\end{equation} 
Finally, the overall loss for RICASSO is:
\begin{equation}
    \label{eq:overall-loss}
    \mathcal{L}_{\text{RICASSO}} = \mathcal{L}_{\text{NOD}} + \lambda_0 \mathcal{L}_{\text{CBCL}} + \lambda_1 \mathcal{L}_{\text{RCL}}
\end{equation}

\section{Experiments}
\label{sec:experiment}

\subsection{Experimental Setups}
\label{sec:experiment-setup}

\paragraph{Datasets}
\label{sec:experiment-dataset}
We compared our long-tailed recognition results on the \textbf{CIFAR10-LT} and \textbf{CIFAR100-LT} datasets \cite{LDAM2019} across different imbalance ratios (IR) of 10, 50, and 100, as well as on \textbf{ImageNet-LT} \cite{ImgNet2009}. It should be noted that these datasets also serve as the ID data in OOD detection.
For OOD detection benchmarking, we use six datasets:
\textbf{SVHN} \cite{SVHN2011}: Contains 26,032 testing images of house numbers, categorized into 10 classes.
\textbf{Texture} \cite{Texture2014}: Comprises 5,640 images of various textures from the Describable Textures Dataset (DTD), with 47 classes.
\textbf{Places365} \cite{Places2017}: Includes 36,000 validation images of scenes and environments, distributed across 365 classes.
\textbf{Tiny ImageNet} \cite{TinyImageNet2015}: A subset of ImageNet, featuring 10,000 validation images across 200 classes.
\textbf{iNaturalist2018} \cite{iNature2018}: Contains 24,426 validation images of various plants and animals, spanning 8,142 classes.
\textbf{LSUN} \cite{LSUN2015}: Features 10,000 validation images across 10 scene categories.

\paragraph{Baselines}
We compare our RICASSO with both conventional long-tailed recognition methods and long-tailed OOD methods, all of which represent the previous state of the art.
For long-tailed recognition, Focal Loss \cite{Focal2017} and LDAM \cite{LDAM2019} are both simple reweighting methods that reweight the loss function in a rebalancing manner. In contrast, RIDE \cite{RIDE2020}, SADE \cite{SADE2022}, and LGLA \cite{LGLA2023} are more complex methods based on multi-expert model. 
For long-tailed OOD methods, we use OpenSample \cite{OpenSampOOD2022}, PASCL \cite{PASCL2022}, Class Prior \cite{CLASSPRIOR2023}, COCL \cite{COCL2024}, BERL \cite{BERL2023}, and EAT \cite{EAT2024}, all of which have already been introduced in Section \ref{sec:related_work}.
Finally, we utilize ODIN \cite{ODIN2018}, a widely-used OOD detector, to perform OOD detection during the testing stage.

\paragraph{Evaluation protocols}
\label{sec:experiment-metrics}
Following the methodologies outlined by \cite{OODMETRIC2021,OODMETRIC2022}, we employ commonly used metrics for evaluating OOD detection and ID classification: (1)  FPR, the false positive rate of OOD examples when the true positive rate of ID examples is 95\% \cite{OODMETRIC2022,OPENMIX2023}; (2) AUROC, the area under the receiver operating characteristic curve for detecting OOD samples; and (3) ACC, the classification accuracy of the ID data. Note that only the average results for OOD detection are reported in main text. 

\paragraph{Implementation details}
Following LGLA \cite{LGLA2023}, we utilized three ResNet-32 expert networks to train the RICASSO framework on the CIFAR10-LT and CIFAR100-LT datasets.
All of the long-tailed recognition baselines are rerun by us under the same settings as ours. Additionally, to ensure fairness for the single-expert network, we maintained the same number of parameters as our multi-expert network. 
For the OOD detection results of iNaturalist2018 on CIFAR-10, we rerun the long-tailed OOD baselines using their settings. However, for the large ID dataset ImageNet, we use the available released weights due to computational resource limitations.
We train the model for 400 epochs, utilizing the SGD optimizer with a momentum of 0.9.
The initial learning rate is set to 0.1, and undergoes a warm-up for the first 5 epochs, during which it is scaled by a factor of 0.1.
Our computational setup includes 8 Nvidia RTX 3090 GPUs and an AMD EPYC 7443 24-core processor.

\begin{table}[tb]
\centering
\small 
\caption{Comparison of long-tailed recognition and OOD detection with previous long-tailed learning methods. Our RICASSO has achieved SOTA in long-tailed recognition and significantly improved OOD detection. The results are presented in the format IR10/IR50/IR100.}
\label{tab:res-LT}
    \begin{tabular}{c|c|c|c|c}
    \hline
    \multirow{2}{*}{Method}     & \multicolumn{2}{c|}{Long-tailed Recognition} & \multicolumn{2}{c}{OOD Detection (CIFAR 10)}
    \\ \cline{2-5} 
                                & CIFAR 10               & CIFAR 100              & AUROC \(\uparrow\)      & FPR95 \(\downarrow\)
    \\ \hline
    Focal Loss \cite{Focal2017} & 90.91/83.10/79.43      & 62.65/50.38/45.02      & 69.44/61.99/61.57& 75.96/87.05/87.98\\
    LDAM+DRW \cite{LDAM2019}    & 89.16/83.10/79.41      & 60.15/48.93/42.67      & 53.80/52.06/52.01& 100.0/100.0/100.0
    \\
    RIDE \cite{RIDE2020}        & 90.18/84.53/80.91      & 64.47/52.38/47.01      & 58.59/55.59/55.31& 95.89/96.57/96.47\\
    SADE \cite{SADE2022}        & {\ul93.03}/89.84/{\ul87.92} & 69.39/58.93/54.12      & 74.43/70.48/{\ul67.94}& 68.43/74.21/{\ul76.43}\\
    LGLA \cite{LGLA2023}        & 92.86/{\ul90.20}/87.80 & {\ul69.88}/{\ul60.60}/{\ul56.50} & {\ul74.28}/{\ul70.75}/66.97& {\ul66.81}/{\ul73.09}/78.22\\
    Ours              & \textbf{93.96}/\textbf{91.30}/\textbf{88.73} & \textbf{71.42}/\textbf{62.35}/\textbf{57.23} & \textbf{93.78}/\textbf{92.06}/\textbf{87.61}& \textbf{22.01}/\textbf{31.04}/\textbf{44.28}\\ \hline
    \end{tabular}
\end{table}

\subsection{Results and Discussion}
\label{sec:results}

\paragraph{Comparison with Current Long-tailed Recognition Methods}
The comparison with long-tailed recognition baselines across CIFAR 10-LT, CIFAR 100-LT, and ImageNet-LT is shown in Table \ref{tab:res-LT} and Table \ref{tab:res-imagenet}. 
In long-tailed recognition for CIFAR 10-LT, our method surpasses the baseline LGLA by 1.1\%, 1.1\%, and 0.93\% for IR10, IR50, and IR100, respectively. For CIFAR 100-LT, our method achieves improvements of 1.54\%, 1.75\%, and 0.73\%, respectively. On ImageNet, our method also demonstrates competitive results with superior OOD detection performance. 
In OOD detection on CIFAR 10, IR 100, as shown in the OOD Detection column of Table \ref{tab:res-LT}, our method significantly outperforms all other long-tailed recognition methods. Compared to our baseline method, RICASSO significantly reduces the FPR95 by 45.68\%, 47.10\%, and 45.26\% for IR10, IR50, and IR100, respectively. In OOD detection on ImageNet, our method also outperforms the baseline by 24.51 in AUROC.
These results affirm the advanced capability of our method in managing the challenges of both long-tailed recognition and OOD detection.

\paragraph{Comparison with Current Long-tailed OOD Methods}
Table \ref{tab:res-ood} provides a comparative analysis of our method against other long-tailed OOD approaches on CIFAR10-LT, IR 100. RICASSO not only achieves the highest long-tailed classification accuracy of 88.73\%, which is 7.17\% higher than COCL, but it also demonstrates competitive AUROC and FPR95. It should be highlighted that all the other methods utilize both ID and real OOD data in their training, while \textbf{we only use ID data}. 
Additionally, a comparison with PASCL \cite{PASCL2022} demonstrated our superiority in OOD detection on ImageNet, as shown in Table \ref{tab:res-imagenet}, rows 1 and 4. Therefore, RICASSO stands out as the most integrated method, mitigating the trade-offs and effectively maintaining the precision in both LT learning and OOD scenarios.

\begin{table}[]
\begin{minipage}[t]{0.48\textwidth}
    \scriptsize
    \caption{Results on ImageNet-LT. We use iNaturalist2018 as OOD dataset.}
    \label{tab:res-imagenet}
    \centering
    \begin{tabular}{cc|c|cc}
    \hline
    Method & OOD? & ACC \(\uparrow\) & AUROC \(\uparrow\) & FPR95 \(\downarrow\) \\ \hline
    PASCL \citep{PASCL2022} & \cmark & 45.50 & 68.90 & 82.06 \\
    SADE \cite{SADE2022} & \xmark & 58.80 & 49.58 & 95.99 \\
    LGLA \cite{LGLA2023} & \xmark & 59.70 & 50.22 & 96.84 \\
    Ours & \xmark & \textbf{59.80} & \textbf{74.73} & \textbf{77.01} \\ \hline
    \end{tabular}
\end{minipage}
\hfill
\begin{minipage}[t]{0.48\textwidth}
    \scriptsize
    \caption{Comparing RICASSO with Other Outlier Exposure Methods on CIFAR10, IR100.}
    \label{tab:res-ood}
    \centering
    \begin{tabular}{cc|c|cc}
    \hline
    Method & OOD? & ACC \(\uparrow\) & AUROC \(\uparrow\) & FPR95 \(\downarrow\) \\ \hline
    PASCL \cite{PASCL2022} & \cmark & 77.62 & 90.69 & 31.10 \\
    COCL \cite{COCL2024} & \cmark & {\ul 81.56} & \textbf{93.01} & {\ul 28.63} \\
    EAT \cite{EAT2024} & \cmark & 81.31 & 89.39 & \textbf{29.12} \\
    Ours & \xmark & \textbf{88.73} & {\ul 90.71} & 37.88 \\ \hline
    \end{tabular}
\end{minipage}
\end{table}


\subsection{Ablation Study}
The ablation study is shown in Table \ref{tab:ablation}. The experiment is conducted on CIFAR 10-LT with a 100 imbalance rate. The OOD metric is the average result on all of the six OOD datasets mentioned in Section \ref{sec:experiment-setup}.

\paragraph{The effectiveness of each module}
As shown in Table \ref{tab:ablation}, incorporating NOD and RCL leads to significant enhancements in all measured metrics (lines 1 and 5). They not only boosts accuracy and FPR for long-tailed classification, but also greatly improves OOD detection efficiency, with an impressive increase of 18.9 in AUROC.
Upon comparing lines 1 and 2, as well as lines 5 and 7, it is evident that AALA enhances the capacity for OOD detection. Although it concurrently impairs LT, this compromise is relatively minor.
A comparison between lines 7 and 8 reveals that CBCL contributes to the performance of RICASSO. While it may slightly impair LT, it enables RICASSO to achieve the optimal OOD performance.

\paragraph{The effectiveness of RICASSO}
The first and last row of Table \ref{tab:ablation} show that, when all components are active, RICASSO demonstrates the highest efficacy in OOD detection while maintaining high accuracy in long-tailed recognition. This highlights the comprehensive effectiveness of all of our proposed techniques.

\begin{table}[]
\caption{The Ablation Study for RICASSO. LTR and OOD refer to the metrics for long-tailed recognition and OOD detection, respectively.}
\label{tab:ablation}
\centering
\small
\begin{tabular}{c|c|c|c|cc|cc}
\hline
\multirow{2}{*}{NOD} & \multirow{2}{*}{RCL} & \multirow{2}{*}{AALA} & \multirow{2}{*}{CBCL} & \multicolumn{2}{c|}{LTR}       & \multicolumn{2}{c}{OOD}
\\ \cline{5-8} 
                     &                      &                       &                       & ACC \(\uparrow\)          & FPR \(\downarrow\)           & AUROC \(\uparrow\)         & FPR95 \(\downarrow\)         \\ \hline
\xmark  & \xmark    & \xmark    & \xmark                     & 87.80          & 13.36          & 69.32& 77.35\\
\xmark                           & \xmark                   & \cmark                     & \xmark                     & 86.85             & 14.62              & 67.63& 77.78\\
\xmark                           & \xmark                   & \xmark                     & \cmark                     & 86.97             & 14.48              & 69.93& 75.88\\
\xmark                           & \xmark                   & \cmark                     & \cmark                     & 86.97         & 14.49          & 68.49& 77.22\\
\cmark                           & \cmark                   & \xmark                     & \xmark                     & { \ul 89.37}             & { \ul 11.81}              & 84.81& 45.93\\
\cmark                           & \cmark                   & \xmark                     & \cmark                     & \textbf{89.40}& \textbf{11.79} & 83.88& 45.93\\
\cmark                           & \cmark                   & \cmark                     & \xmark                     & 88.87   & 12.38          & { \ul 89.42}& { \ul 41.77}\\
\cmark                           & \cmark                   & \cmark                     & \cmark                     & 88.73         & 12.36    & \textbf{92.97}& \textbf{32.77}\\ \hline
\end{tabular}
\end{table}

\section{Conclusion}
In this paper, we propose a unified framework, RICASSO, for long-tailed recognition and OOD detection. We are the first to eliminate the need for OOD data in long-tailed OOD recognition tasks.
We achieve this by replacing OOD data with mixed data, which is used in three ways: firstly, to conduct pseudo outlier exposure; secondly, to help the model better discriminate between head and tail classes; and thirdly, to serve as a virtual boundary in contrastive learning, enhancing clustering in feature space. By doing so, RICASSO successfully mitigates the trade-off between these two tasks, achieving state-of-the-art performance in long-tailed learning and significantly improving OOD detection compared to our baseline method. Extensive research shows that RICASSO exhibits the best comprehensive performance in long-tailed OOD recognition.



\bibliographystyle{unsrtnat}
\bibliography{neurips_2024.bib}

\end{document}